\documentclass[11pt]{article}
\usepackage{geometry}
\usepackage{graphicx}
\usepackage{amssymb}

\title{Quantum Reasoning using Lie Algebra for Everyday Life (and AI perhaps\ldots)}
\author{Steven Gratton\\\texttt{stg20@cam.ac.uk}\\Kavli Institute for Cosmology Cambridge, Institute of Astronomy, 
\\University of Cambridge, Madingley Road, Cambridge, CB3 0HA, U.K. }
\date{November 6, 2018}

\def\({\left(}
\def\){\right)}
\def\[{\left[}
\def\]{\right]}
\def\<{\left<}
\def\>{\right>}

\newcommand{\ba}{\begin{eqnarray}}
\newcommand{\ea}{\end{eqnarray}}
\newcommand{\labeq}[1] {\label{eq:#1}}
\newcommand{\refeq}[1] {(\ref{eq:#1})}

\newcommand{\labsec}[1] {\label{sec:#1}}
\newcommand{\refsec}[1] {(\ref{sec:#1})}

\begin{document}
\maketitle

\begin{abstract}
We investigate the applicability of the formalism of quantum mechanics
to everyday life.  It seems to be directly relevant for situations in which the very
act of coming to a conclusion or decision on one issue affects one's confidence
about conclusions or decisions on another issue.  Lie algebra theory is argued to be
a very useful tool in guiding the construction of quantum descriptions of such situations.
Tests, extensions and speculative applications and implications, including for the 
encoding of thoughts in neural networks, are discussed.  It is suggested that the recognition
and incorporation of such mathematical structure into machine learning and artificial intelligence might lead 
to significant efficiency and generality gains in addition to ensuring probabilistic
reasoning at a fundamental level.

\end{abstract}

\section{Introduction}

Bayesian inference, as espoused by e.g.\ Jaynes in \cite{jaynes}, is the logical way to associate
real numbers with plausibilities of propositions.  Quantum mechanics, on the other
hand, requires complex numbers on its way to prediction about the
physical world.  Many have speculated on what quantum theory is telling us about
reality, see e.g.\ Penrose \cite{enm}, Deutsch \cite{fabrel} and Aaronson \cite{qcsd}, and whether it can be thought
of as an improvement or extension of probability theory/bayesian inference in some way,
see e.g.\ \cite{Deutsch:1999gs,Barnum:1999ew,2002PhRvA..65b2305C}.

This paper takes the point of view that the structure of quantum mechanics is 
indeed an extension to bayesian inference.  As such, there should be ``real-world"
situations for which the quantum formalism provides a good way to describe
them.  We actively seek out some examples and then go on to draw some implications.

One of the hallmark features of quantum mechanics is the Heisenberg Uncertainty 
Principle, expressing the fact that measuring some property of a physical system
typically disturbs the system.  Another is the fact that certain variables are ``quantized", or 
can only ever be measured to have one of a discrete set of values. 

The first feature finds expression in quantum mechanics by the representation of 
dynamical variables by matrices that do not necessarily commute with each other.  The second feature 
finds expression by the result of measuring the dynamical variable only being an eigenvalue of
the corresponding matrix, of which there are a finite number (when the matrix is finite-dimensional).  

Thus we expect quantum reasoning in the everyday world will be of benefit for situations
in which coming to a conclusion about some thing affects one's conclusions about other things
and in which only a limited number of outcomes are possible.

In related work, mathematical psychologists have speculated whether the brain might be well-modelled as 
reasoning in a quantum fashion; see \cite{bb2012} for an introduction to this field.   Our point of departure
here is with the recognition of the potential of Lie algebra theory, see e.g.\ \cite{georgi}, to act as an ordering 
principle in the construction of quantum descriptions.

In physics the appreciation of symmetry has led, to take just two examples, to the Standard
Model description of nature (see e.g.\ \cite{donoghue}), in which particles are grouped into families and
 are precisely related to one another, and to the celebrated 
Wigner-Eckart Theorem (see e.g.\ \cite{edmonds}), which 
allows one to relate whole classes of expectation values of quantities between pairs of states to one another
by rotational symmetry alone. 

To set the scene, we first discuss a classic experiment demonstrating the necessity of a 
quantum description of nature.  Next, we review the mathematics of quantum theory and 
of Lie algebras.  Then, we walk through a simple example involving choosing a drink at a
restaurant to show how the formalism might all apply.  After this we discuss simple tests of
and extensions to the scheme, and then we close with some hopefully-entertaining speculation on 
what these kinds of ideas might imply for the understanding of thought and for the development
of machine learning and artificial intelligence.

\section{\labsec{sterngal} Lessons from Stern-Gerlach}

This section discusses an archetypal example of a quantum mechanical behaviour in the
physical world.  Our discussion stresses more than most the effects of one measurement
on the results of the next measurement, and might be of primary interest to physicist readers.
For non-physicists, hopefully enough of a flavour of the rationale for quantum mechanics
will come through to enable them to proceed with the rest of the paper.

Many quantum physics textbooks (e.g.\ \cite{jjs} on which this discussion is partially based) include discussions of ``Stern-Gerlach"
experiments.  The quantum system is a single particle that has some ``spin" associated with it (think of a spinning top
whose angular motion is described by an axis and a rotational frequency).  It is described how one can measure the 
particle's ``spin" in some direction
in space, by passing the particle briefly through a magnetic field that is inhomogeneous (changes in strength)
in the direction in question.  This changes the momentum of the particle along that direction by an
amount that is proportional to its spin. So if we say the particle starts off travelling in the
z-direction, and we want to know what it's spin is in the x-direction, we allow it to pass
through a magnetic field that changes in the x-direction.  For a particular sort of particle, a ``spin-1/2'' one,
the particle will either get a nudge of a specific amount in the x-direction or a nudge of the same amplitude 
but in the opposite, negative-x-direction.  The nudge should be proportional to the spin of the 
particle.   
This shows that the spin of the particle along the x-direction can only take one of two
discrete values, and hence is quantized.  Similarly, if one repeats
the experiment with the magnetic field rotated to change in the y-direction, one finds the spin in the
y-direction is similarly quantized.  
  
One can next consider two measurements, one after the other, by adding a second magnetic field along each
nudged path.
Align both fields along each path to be inhomogeneous in the x-direction say.  If the particle got a nudge in the positive-x direction
from the first magnet, it gets another nudge in the positive-x direction from the second magnet.  Likewise,
if it got a nudge in the negative-x direction, it gets another nudge that way.  So, repeating the same
measurement yields the same result.   

Next though, rotate the second magnetic field along each path to measure the spin in the y-direction with a new magnetic field.  
It now emerges with either a nudge in positive-y direction or in the negative-y direction.  If this two-step experiment is repeated many times, the y-nudge
appears to be statistically independent from the x-nudge.  

One may now add a third magnetic field along each path.  Let the first and third magnets measure the spin in the x-direction, but the second
measure the spin in the y-direction.  Repeating the experiment many times, the last x-nudge appears to be statistically independent
of both the preceding y-nudge and the first x-nudge!  Indeterminacy appears to have entered physics and the Uncertainty Principle
is well illustrated, in this case between the x and y components of the particle's spin.  

Faced with these facts, one often breaks the quantum description of the particle's history into stages, three in this case, ``streamlining" the description
immediately after each measurement to one in which the result was certain to have had emerged.  By fiat then, the next spin measurement 
becomes independent of results earlier than the immediately preceding one.  

But this is not necessary.  Von Neumann \cite{vneu} showed how a quantum description of the system might be maintained by treating 
each measurement quantum-mechanically rather than as a ``jump" to a streamlined state.  Here, the key fact is that electromagnetism
provides for an interaction between the spin of a particle and the magnetic field it is in.  A changing magnetic field gives the nudge
discussed above, the stronger the gradient the stronger the nudge.  We need a sufficiently strong field in order to adequately
separate the two beams.  The required strength is set by first imagining turning the field to zero, when there would be no systematic nudge, whilst still
keeping the apparatus in place. There would still be a scatter
of the particle's motion, as though a random nudge centred around zero
with some width along the measurement direction was now applied, however.  This
is a residual effect of the apparatus, it now just acting as a finite-sized ``aperture'' or hole that the particle has to pass through.  This
spread is reminiscent of the diffraction of a sound wave through a door, and indeed is the canonical illustration of both the ``wave-particle"
duality of quantum mechanics and of the Heisenberg Uncertainty Principle: constraining a particle's position in some direction necessarily
incurs a spread in the ``conjugate momentum" (proportional to its velocity) in that same direction.  The smaller the aperture, the larger the spread.

With the field turned back on, there is both the systematic nudge parallel or antiparallel to the field direction on top of the diffractive spread.  
We need the field gradient to be strong enough so that the systematic effect is larger than the spread in order to be confident of the
spin value and so effect a proper measurement of the spin (see e.g.\ \cite{bohm}).  There is thus a minimal change in field strength
from one side of aperture to the other consistent with performing a spin measurement.

Now, electromagnetism tells us that a magnetic field has another
effect, this time on the spin itself of a particle: it causes the spin to \textit{precess}, or rotate,
around the magnetic field direction, just as the axis of a spinning gyroscope rotates slowly around the earth's gravitational field direction.  
For a given time of exposure to the field, the total angle of rotation is proportional to the field strength.  

We are now finally in a position to see just how measuring the spin in the y-direction affects the spin in the x-direction in between two
x-measurements: as well as nudging the particle, the y-magnet also causes the spin to precess around the y-direction, wanting to change
x-spin into z-spin and back whilst keeping the y-spin unchanged.  But
because the field strength has to change across the aperture in order
to effect a spin measurement, the
amount of rotation is not fixed, depending in a sense on the unknown details of the particle's path through the aperture.  Indeed, it turns 
out that the minimal change in field strength across the aperture required for a y-spin measurement is sufficient to lead to at least a
half-turn variation in the amount of spin rotation in the xz-plane, rendering the x component of the spin unpredictable.  

The interplay between spin measurements, considered quantum mechanically throughout, is consistent with Heisenberg's Uncertainty 
Principle applied directly to the spin components, as one would expect.  However, it is important to review the quantum description of the
measurement process as we have just done in order to understand that the uncertainty caused by ``measurement'' indeed has a physical basis.  This allows us to
view the ``streamlining'' procedure, where the quantum effects of the measuring systems are not explicitly taken into account but rather the quantum
evolution of the system in question is restarted afresh after each measurement, as an approximation
to the full quantum description.  The difficulty of course is then in seeing how probabilities come into the full quantum picture, an issue to which we
shall return at the end of the next section.

\section{\labsec{olagt} Operators, Lie Algebras and Gleason's Theorem}

From the above reflections on the nature of measurement from a quantum perspective, we can see perhaps why the results of measurements
have to be eigenvalues.  A ``measurement'' is a stable correlation of the property in question with something else, caused by some interaction.  
The interaction thus has to involve the quantity being measured, and hence different eigenvalues of the quantity correlate in different ways.

Let us now see how this helps our main business of improving inference in everyday life.  From quantum mechanics we carry over the idea
of representing questions or propositions by matrices.  In physics we have seen that making a measurement corresponds to applying an
interaction to a system, changing it in certain ways.  So we expect the formalism to be at its most useful in situations where pushing for an 
answer to one question can change the answer to another one.  A moment's reflection suggests such situations are not too uncommon:

\begin{itemize}
\item getting one's children to accept a common meal for dinner: does offering steak affect the
chances of them all accepting fish?
\item deciding on a family day trip
\item a person contemplating whom to ask on a date: does the very act of deciding to come to a decision to ask
A force him/her to reevaluate his/her feelings for B?
\item a jury coming to a common decision: would coming to a conclusion
on point B after coming to a conclusion on point A reopen the discussion on point A?  
\item an advisory committee coming to a common position on one of a number of questions posed to it
\item a government choosing how to respond to some provocation: war, sanctions or appeasement?
\item a negotiator trying to model how another entity might respond to different approaches 
\item an interrogator considering how to extract information from a subject
\item tax/trade policies on economies
\item an investor deciding which start-up to invest in
\end{itemize}

Many of these examples involve inter-entity communication.  It is also interesting that some examples are introspective, involving
uncertainty within the questioner, whereas others involve modelling external uncertainties.  Some examples are passive, merely
wanting to correctly understand a situation, whereas others are more active, suggesting courses of action (or influences on courses
of action).  

For a given problem we
shall imagine we are posed with a number of different ``questions" or ``options''.  Depending on the situation in hand, some of the 
questions may be related to one another.  For example,
for an adult, broccoli, peas and sweetcorn might be a truly independent choice of vegetables to consider, whereas for a young 
child, having peas might be considered just a kind of intermediate case between having broccoli or sweetcorn.  Or, a selection committee for 
an academic might find itself effectively evaluating each candidate by how much he or she fits in to one of a few existing groups within a department and 
going from there.   We shall denote the underlying dimensionality of the ``option space'' by $d$.   

To be able to use mathematics we need to associate
numbers with potential answers or outcomes: typically zero might correspond to ``no preference", a large positive number to ``strong agreement'', 
a small negative number to ``weak disagreement" and so on.   Different people, groups or electorates might be able to return answers
varying in subtlety to the same sets of options; possible results might come in different \textit{representations}.  

So, we want to use the apparatus of quantum mechanics to address $d$ ``entwined'' options.  Let us denote the options $t_i$, $i=1,\ldots,d$.  
In the early editions of his seminal book on quantum mechanics \cite{dirac}, Dirac showed how one could ``build up'' the quantum description
of a spinning particle simply from assuming the so-called commutation relations between the operators $\mathbf{x}$, $\mathbf{y}$ and $\mathbf{z}$ 
representing the x-, y- and z- spin components:
\ba
\mathbf{x} \mathbf{y} -\mathbf{y} \mathbf{x} &=& i \mathbf{z}, \nonumber \\
\mathbf{y} \mathbf{z} -\mathbf{z} \mathbf{y} &=& i \mathbf{x}, \nonumber \\
\mathbf{z} \mathbf{x} -\mathbf{x} \mathbf{z} &=& i \mathbf{y}. \labeq{comrel}
\ea
In quantum mechanics it is shown that operators that do not commute cannot be measured simultaneously.  
So the fact that no two components of the spin commute with each other (i.e. the right hand sides are nonzero) corresponds to the fact 
that the spin-components cannot be measured simultaneously.
For this example, no talk of dynamics is necessary, so it forms a particularly suitable starting point for our discussion.  Dirac shows that $\mathbf{x}^2 + \mathbf{y}^2 + \mathbf{z}^2$ (now called the quadratic Casimir operator) \textit{does} commute with each of $\mathbf{x}$, $\mathbf{y}$ and $\mathbf{z}$.  Treating the operators as square matrices, one is
able to find two-by-two triples, three-by-three triples and so on to arbitrarily large dimension of matrices that satisfy the commutation relations \refeq{comrel} and in which the Casimir operator is proportional to the identity.  The multiple of the Casimir over the identity increases with the dimensionality $d_r$ of the representation (one using $d_r \times d_r$ matrices).  

One might be puzzled about the non-uniqueness in the choice of
matrices for a representation of given dimensionality: it turns out
for this problem that different choices are \emph{equivalent}, meaning
that each of one set is given by pre- and post- multiplying the
corresponding one of the other set by some other matrix $U$ and its inverse.  By inspection though one can see that this preserves the commutation relations \refeq{comrel}, and matrices related in this manner have the same eigenvalues and so give the same spectrum of potential results for a measurement.  (Later on we shall see an example in which \emph{inequivalent}
representations do occur.)  One can use this freedom to choose a representation in which one of the three matrices representing the operators is diagonal; we can then just read off its 
eigenvalues.  (The \emph{rank} is the number of operators that can be simultaneously diagonalized, so here the rank is one.)

Note that one can find sets of matrices of some dimension that satisfy the commutation relations but whose Casimir operator is not represented by a multiple of the identity matrix.  Such representations
are called \emph{reducible} though, since they can be effectively ``built up'' from combinations of the \emph{irreducible} representations just described.  The spin-1/2 particle discussed in Sec.~\refsec{sterngal} above is described by the 2-dimensional irreducible representation.  A pair of such particles could be described by the product of two such representations.  This is reducible however
into the sum of a one-dimensional and a three-dimensional representation.  (The one-dimensional representation may be thought of as a special case in which all operators are represented by zero, a trivial solution of \refeq{comrel}).  

So far we have focussed on the representation of the operators.  Now let us turn to the states the particles might be in. 
A given quantum spin state is represented by a column vector of (generally complex-) numbers.  The standard rules of quantum mechanics (see Dirac~\cite{dirac})
tell us how to extract probabilities for measuring results from the column vector.  We return to the justification of these rules below. 

It may seem strange at first that both \textit{prima facie} different questions are related to one another and that different states determining possible answers are related to one another.  Let us spend a few moments more on this
point.  In free space, no set of axes is preferred over any other (an aspect of Lorentz invariance).  So, rather than choosing to use  $\mathrm{x}$, $\mathrm{y}$ and $\mathrm{z}$ as our base axes,
we might for example choose axes $\mathrm{x}'$,
$\mathrm{y}'$ and $\mathrm{z}$, rotated a few degrees
around the $\mathrm{z}$ axis from our old ones.  The operators
corresponding to measurements in the new spin directions will be
linear combinations of the old operators.  (Physical rotation
invariance is encoded by the conservation of the commutation relations
under the rotation.)  Indeed, the operator for measuring the spin
component in \emph{any} direction will be a linear combination of the
$\mathbf{x}$, $\mathbf{y}$ and $\mathbf{z}$ operators.  So questions are certainly decomposable into others, and the dimension of the question
space may well be finite, even though a continuous infinity of
different questions might be askable (corresponding in this example to
all the directions in space along which spin may be measured).  Now
consider the answers a state represents.  If the state represents certainty of a particular answer to a particular question, the state vector will be the eigenvector corresponding to the particular answer or eigenvalue of the matrix representing the question.  An eigenvector of one matrix is not normally an eigenvector of another matrix that does not commute with the first.  So the state does not normally correspond to a definite answer to a second question also.  However, an eigenvector of the first matrix can be expressed as a complex-number-weighted sum of eigenvectors of the second, and vice versa.
So, for the spin-1/2 particle of \refsec{sterngal}, spinning in the $+$y direction is a particular superposition of spinning in the $-$x and $+$x directions and so on.  Indeed, it is worth bearing in mind that Dirac stressed that the \emph{superposition principle} is one of the great simplifications of quantum mechanics over classical mechanics and that it is actually \emph{allowed in} by indeterminacy.

The example above, emerging remarkably out of Eqs.~\refeq{comrel}, seems to set an ideal precedent for what we would like to do for our everyday inference and decision making.  However, it only can
be applicable for cases with dimensionality $d=3$.  Fortunately, Eqs.~\refeq{comrel} simply define a particular case of what is generally known as a \emph{Lie algebra}, the mathematical understandings of which are mature.  Indeed, the description above has been purposely shaped with an eye to moving to a general \emph{compact simple} Lie algebra.   The technical conditions of compactness and simplicity
ensure that suitable finite dimensional matrix representations can be found (see e.g.\ \cite{weinberg2}).  Compactness basically corresponds to only having a ``finite-volume"'s worth of questions to consider (in the example given above, think of the sphere's worth of potential spin measurements as opposed to say an infinite plane's worth of measurement options).  Simplicity roughly means that the algebra has already
been ``broken down'' as much as possible.  One can then consider other Lie algebras by ``summing up" combinations of compact simple ones (along with ``U(1)'' Lie algebras that we have not discussed here).   

The compact simple Lie algebras were completely classified by the end of the 19th Century.  
There are four general families, each of them parametrized by an integer $n$, denoted $a_n$, $b_n$, $c_n$ and $d_n$ (or alternatively su($n+1$), o($2n+1$), usp($2n$) and o($2n$)), with only a few isomorphisms or overlaps ($a_1=b_1=c_1$, $c_2=b_2$, and $a_3=d_3$). One needs
$n \geq 1$  for the first three, and $n \geq 3$ for the last. $n$ gives the rank of a representation, and the dimensions $d$ are given by $n(n+2)$, $n(2n+1)$, $n(2n+1)$ and $n(2n-1)$ respectively. There are also five ``exceptional" cases, denoted $g_2$, $f_4$, $e_6$, $e_7$ and $e_8$, the subscript indicating the rank, and these have dimension 14, 52, 78, 133 and 248 respectively.   
Generalizing \refeq{comrel}, the operators or generators $t_a$ can be chosen to satisfy:
\ba
[t_a,t_b]=i f_{abc} t_c
 \labeq{tcom}
\ea
where $f_{abc}$ are called the \emph{structure constants} of the algebra.  The $t_a$'s may be taken to be Hermitian traceless matrices (often conventionally normalized to satisfy $\mathrm{tr} \(t_a t_b\)=\delta_{ab}$/2), the commutator $[t_a,t_b]$ is shorthand for $t_a t_b - t_b t_a$, and repeated indices are implicitly summed over.  The structure constants are real and are antisymmetric in all three indices. They are constrained by the Jacobi identity,
\ba
[[t_a,t_b],t_c]+[[t_b,t_c],t_a]+[[t_c,t_a],t_b]=0
\labeq{jacobi}
\ea
that must hold by virtue of the definition of the commutator and the associativity of matrix multiplication. 

The structure constants define a $d_r=d$-dimensional representation, called the \emph{adjoint} representation, with 
\ba
 (t^\mathrm{A}_a)_{bc}=-i f_{abc}
 \ea
 (i.e. the bc'th component of the a'th matrix is given by $-i$ times
 the indicated structure constant).  The adjoint representation is not the only one that exists, and is not usually the smallest either.  In fact one may construct ``fundamental'' representations and then ``build up'' higher dimensional representations
from the fundamental one(s), see e.g.\ \cite{georgi,ramond}, so one can label representations by the numbers of each of the fundamental representations they involve.  One may construct higher-order analogues $C_3$, $C_4$,~\ldots\ of the quadratic Casimir operator $C_2$ and use suitable sets of these values to label representations also.

Given one set of matrices forming a representation of \refeq{tcom}, one can form another set of matrices by taking the
negative complex conjugate of each element of each of them, since  \refeq{tcom} implies:
\ba
[(-t^*_a),(-t^*_b)]=i f_{abc} (-t^*_c).
\ea
The new matrices may or may not form an equivalent representation, in the sense described above, to the original set.  As an equivalent
matrix has the same eigenvalues as the original one, if any matrix in the original set has eigenvalues that are not symmetric around zero,
then the two representations must be inequivalent. 

Now that we have reviewed all the Lie theory we shall need, let us now turn back to the question of interpretation. In quantum mechanics proper
there are multiple arguments for the standard rules for obtaining
probabilities from the formalism, e.g.\
\cite{dirac,Deutsch:1999gs,Hardy:2001jk}.  There is the issue though
that a probabilistic theory could be describing an \emph{approximation} to a reality in which probabilities do not fundamentally appear, e.g.\ \cite{fabrel}.  For our application though this difficulty 
does not apply since our whole purpose is to calculate probabilities.  Then, given we want the Lie algebra-type structure and to represent questions by matrices,
perhaps the most compelling argument comes from Gleason's Theorem.  This theorem tells us that the standard scheme is basically the only one way to consistently link probabilities
to operators, at least in a representation of dimension $d_r > 2$ and as long as we want the probability for a given outcome of any question
to be independent of ``rotations'' between the other outcomes we might have received.  
The standard scheme of course works
for $d_r=2$ and we continue to use it in this case, even if it is not still proven to be unique.  Strictly, Gleason's Theorem shows that a ``density matrix'' description
of the system must exist, which can be thought of as a classical probability distribution over pure quantum states (see e.g.\ \cite{shadmind}).  In this work we shall
assume we can describe a system by a pure state.  See the discussion in \cite{Barnum:1999ew} for a forceful advocation of the role of Gleason's Theorem in deriving probability
in quantum mechanics.   

In conclusion, the framework of identifying matrix generators of Lie algebra representations with questions or options and then modelling a system and extracting probabilities
from it just as in quantum mechanics seems to be a secure way to reason about everyday life.  We now turn to an example treatment of a situation in this maner.

\section{A Drink, Sir/Madam?}

Imagine you are a restaurant owner, and wish to serve clients drinks. Knowing your varying stock levels, 
profit margins or simply for speed, you may wish to ``steer'' each of the diners towards certain options, simply by asking each of them 
if they would like a particular drink rather than asking them for a general preference.  

Hence it would be helpful to have a model in mind of how people might respond to being asked about a drink.  Children might be
expected to have a simpler taste and so need a lower dimensional ``space'' of drink possibilities compared to an adult (i.e.\ a lower $d$ for children).  They might
also have a simple like/dislike response whereas adults might express more subtle preferences.    So we expect children to have
a lower dimensional representation space of answers than for adults (i.e.\ a lower $d_r$ for children).  Children might only have one drink in mind they'd definitely
accept, whereas adults might have two or more drinks they'd definitely accept (i.e.\ a lower rank $n$ for children).  Reflecting on one's own experiences when out for dinner, 
and looking at the differences in the child and adult menu options, might attest to the relevance of these considerations!

The rank $n$ and dimensionality $d$ of potential algebras and the
dimensionality $d_r$ of potential representations are often helpful to
keep in mind when determining 
which algebra to try.  For a single child, we shall consider su(2), and for a single adult we shall consider su(3).  

\subsection{su(2)}

This algebra has three generators, which for illustration we can take to be:
\begin{table}[!h]
\centering
\begin{tabular}{c c c}
$t_1$: cola? & $t_2$: apple juice? & $t_3$: water? \\
\end{tabular}
\end{table}

\noindent Asking a child about a different drink, e.g.\ lemonade, should then be thought of as a superposition of asking him or her about these three drinks, say between
cola and water.  As the rank is one, such a child can have a definite opinion about only one drink at a time.  

Now let us consider which representation to use.  The lowest nontrivial representation has $d_r=2$, so let us consider that one first.  Responses can only be for/against.
The superposition principle tells us that a child who'd definitely have cola may or may not have water with equal probability. 

Now imagine we have two children, trying to decide on a jug of drink to share.  If each of them are in the $d_r=2$ representation, then combined they can either disagree 
about any option or agree strongly about an option.  Indeed, a product
of two $d_r=2$ representations reduces to the sum of the one-dimensional and three-dimensional irreducible ones.
  If a younger child and an older child are together, the older child might offer a stronger 
but more refined view (strong like/no preference/strong dislike).  In
this case, the product of a $d_r=2$ and a $d_r=3$ representation turn
out to combine to the sum of a $d_r=2$ one (weak preference)
and a $d_r=4$ one.  Here we have assumed the children are asked independently.  If however the younger child is scared of the older one, he or she might not respond and the correct 
description of the combined system might just be that of the older child alone.  

\subsection{su(3)}

Now let us consider adults in su(3).  This algebra has eight generators and has rank two.  Conventionally $t_3$ and $t_8$ are chosen to commute, and the nonzero structure constants
are set by $f_{123}=1$, $f_{147}=f_{246}=f_{257}=f_{345}=1/2$, $f_{156}=f_{367}=-1/2$ and  $f_{458}=f_{678}=\sqrt{3}/2$.  Diners will often accept water 
without prejudicing a decision to also have an alcoholic drink for example, so we reflect this in our association of generators with questions:\\
\begin{table}[!h]
\centering
\begin{tabular}{l l l l}
$t_1$: wine?
&$t_2$: whisky?
&$t_3$: beer?
&$t_4$: coffee? \\
$t_5$: tea?
&$t_6$: lemonade?
&$t_7$: cola?
&$t_8$: water? \\
\end{tabular}
\end{table}

\noindent Asking about champagne might be linear combination of asking about lemonade and wine, mainly along the wine direction say.  So if the restaurateur overhears a person expressing a desire
for a glass of champagne perhaps but the last bottle has already gone, he or she might do well to ask that person about a glass of wine.  For certain drinkers, questioning about 
lager might be the negative of questioning about beer.

su(3) is more subtle than su(2) in that it has representations in only certain dimensions, and has inequivalent representations for certain of those dimensions too.  Indeed, its lowest nontrivial
representations are in dimension three, and there are two inequivalent ones.  These are typically denoted by 3 and $\bar{3}$, and the (single) eight dimensional representation is the adjoint one 
denoted 8.  The Casimir operator $C_2=4/3$ for 3 and  $\bar{3}$, and  $C_2=3$ for 8, indicating the increasing range of feeling expressible as one goes to higher dimension representations.  In the 3 representation,
any state can be expressed as a combination of the simultaneous eigenvectors of $t_3$ and $t_8$ with the following combinations of eigenvalues:\\
\begin{table}[!h]
\centering
\begin{tabular}{c c c c}
$t_3:$&$-1/2$&0&$1/2$\\
$t_8:$&$1/(2\sqrt{3})$&$-2/(2\sqrt{3})$&$1/(2\sqrt{3})$.\\
\end{tabular}
\end{table}

\noindent These eigenvectors also have definite values of $I^2\equiv t_1^2+t_2^2+t_3^2$ of $3/4$, $0$ and $3/4$ respectively. The other $t$'s can have the eigenvalues $-1/2$, $0$ or $1/2$.

In the $\bar{3}$ representation, 
any state can be expressed as a combination of the simultaneous eigenvectors of $t_3$ and $t_8$ with the following combinations of eigenvalues:\\
\begin{table}[!h]
\centering
\begin{tabular}{c c c c}
$t_3:$&$-1/2$&0&$1/2$\\
$t_8:$&$-1/(2\sqrt{3})$&$2/(2\sqrt{3})$&$-1/(2\sqrt{3})$.\\
\end{tabular}
\end{table}

\noindent Again, these eigenvectors also have definite values of $I^2$ of $3/4$, $0$ and $3/4$ respectively, and the other $t$'s can have the eigenvalues $-1/2$, $0$ or $1/2$.

We see the asymmetry of the allowable strengths of preferences for and against water in the two representations, and how this couples to any desire for alcohol (given by $I^2$).

The adjoint representation by contrast is more democratic: \\
\begin{table}[!h]
\centering
\begin{tabular}{c c c c c c c c c}
$t_3:$&$-1$&$-1/2$&$-1/2$&0&0&$1/2$&$1/2$&$1$\\
$t_8:$ & 0 & $-\sqrt{3}/2$ & $\sqrt{3}/2$ & 0 & 0 & $-\sqrt{3}/2$ & $\sqrt{3}/2$ & 0 \\
\end{tabular}
\end{table}

\noindent with $I^2$'s of $2$, $3/4$, $3/4$, $0$, $3/4$, $3/4$, $3/4$, and $2$ respectively (the value of $I^2$ happening to distinguish the two eigenvectors with the same eigenvalues for $t_3$ and $t_8$).  
The other $t$'s share the same eigenvalues as $t_3$ (but not simultaneously).  

Combinations of desires now join interestingly also.  For example, a couple consisting of two 3's would reduce to the sum of the 6 representation and a $\bar{3}$ representation, whereas a 3 and a $\bar{3}$ 
would reduce to the sum of the adjoint and the one-dimensional representation.   
If a couple's preferences both fall into the adjoint representation, their independent summed responses are reducible into the sum of the 27 representation, the 10 and the $\overline{10}$ representations, two copies of the adjoint representation and the one-dimensional representation.  

Of course, the pair in a couple might take each other's preference into account
before responding jointly to the restaurateur, and so their combined
response might fall into a single representation. 

By considering all these factors, a restaurateur should be able to optimize his or her drinks offerings for both adult and child clients in a number of situations.  One interesting
problem would be how to model thinking about a common drink for a family, if members of the family would not individually fall into the same algebra.  Perhaps the adults
instinctively ``drop down'' to the children's level.  Restaurateurs also might have some experience in how people respond if they have chosen one drink which then turns
out to be unavailable.  Does the second response indeed follow the quantum probability predictions?

It would be interesting and should be possible to perform an analysis on restaurant receipts to investigate at least some of the questions, particularly if one would
be prepared to do tests on differing ways of presenting options to customers at self-serve fast food ordering points for example.    

\section{Further Work}

\subsection{Testing the Predictions}

As just suggested, there is plenty of scope for testing the predictions of this approach, and there might exist plenty of ``big data'' sets that confirm such behaviour
in certain situations, if only one can identify the appropriate questions to consider together.

In the above example, a simple observation is that many set menus for children offer a choice of three meals, three desserts and three drinks, presumably
trying to span the space of acceptability for most children in the
minimal way (the author's children certainly often behave as spin-1/2
particles with regards to their culinary preferences\footnote{A
  humorous puzzle for physicist parents: can children somehow manage to disagree on dinner even whilst being spacelike separated and thus violate the Bell inequalities? Hidden variables or superluminal communication perhaps?!}).

A very basic consistency check is that if the exact same question is simply repeated (for example if the answer was inaudible), one should expect to receive the same answer.  This is what the 
standard quantum formalism gives, the state of the system after measurement being given by the eigenvector corresponding to the eigenvalue representing the measurement result.

\subsection{Extending the Maths}

The mapping of preferences onto specific real numbers (the eigenvalues of a matrix representing the question) doesn't 
always feel totally natural: things seem less clear cut for relative strengths of feelings say than for spin components.  
For a
given system, perhaps one should 
view such a mapping merely as a device, much as with the relation between ``plausibilities'' and ``probabilities'' in the construction 
of Bayesian inference \cite{jaynes}.  When systems are ``combined'' however it seems important that both have the same ``scale''
of interpretation, in order say that a like and dislike precisely
cancel.  

Such discussions might remind one of ``symmetry breaking''
in particle physics, and it would be natural to search for analogous situations in everyday life.  Indeed particle physics with its ``eightfold
way'',  uncovering the importance of su(3) in understanding hadron spectroscopy \cite{eightfold}, was a 
motivating inspiration for the scheme presented here.     

Above we argued that Lie algebras seem the perfect fit for our need of generating matrices to represent options that
should be considered together and can be combined or rotated into new ones.    Generalizations of Lie algebras do exist
and are useful in physics, for example Lie \emph{superalgebras} as used in supersymmetry.  It would be interesting
to see for which if any situations a superalgebra treatment might be appropriate.

It is not always immediately clear how to interpret certain outcomes.  In the child's drink example above, consider the negative 
of the matrix representing a question about water.  Can this just be a query about ``not-water'', or does it have to be a query about 
another drink, orange juice say? 

If for some situation only a fixed set of questions makes sense, rather than a continuous array of possibilities, perhaps finite groups and
their representations can replace Lie algebras and their representations.  Indeed, finite simple groups have recently also been completely classified, 
and just as for Lie algebras there are a number of countable infinite families of them along with a collection of ``sporadic'' ones. 

\subsection{Applications and Implications}

It might behove one to respect, rather than to possibly unwittingly try and fight, an underlying Lie algebra structure to a situation
if indeed it is there.   This would involve 
ensuring that one frame one's problems in an appropriate dimension, 
adding an imaginary company or two to an accelerator investment
discussion for example, or considering some candidates as linear combinations of 
others if that effectively represents a job search committee's inner workings for example.  It would also ensure making sure one chooses an
appropriate representation, particularly if there are inequivalent ones with the the same $d_r$.  
Menu or questionnaire design could clearly be influenced.  For example, Likert-type scales \cite{likert}, measuring attitudes in some discrete manner 
(e.g.\ ``1=strongly disagree \ldots\ 5=strongly agree''), seem to correspond well with the discrete eigenvalues of finite-dimensional operators.  
But when two (or more) questions are asked, certain algebras and representations suggest one should not necessarily use the ``direct product'' of 
such one-dimensional scales but might rather need a higher-dimensional lattice of responses, even when the questions commute (consider the allowed
$t_3$ and $t_8$ responses in su(3) as discussed above).  In fact, restricting possible outputs/responses might be considered as a device that compels a
person/system to fit into some representation.

Perhaps people should be able to vote either for or also \emph{against} a candidate of their choice in elections, more closely corresponding
to the su(2) structure of a two-way choice (unless a preference against one candidate happens to be identical to a preference for another one).
It would be fascinating to do research on what
sorts of effects this would have, perhaps leading to a more satisfied electorate with an easier way to remove bad leaders 
and so to correct errors \cite{begin}. 

Recent work on machine translation using neural networks, building on earlier ideas of representing words in a vector space \cite{2013arXiv1301.3781M}, has suggested that it is profitable to compress sentences expressed in one language
into ``sentence vectors'', which can then be decompressed into another language \cite{SutskeverVL14}.  This suggests that thoughts themselves may occupy a sort of vector
space in $O(1000)$ dimensions \cite{hinton}.  It would be fascinating to see if these insights from deep learning can be combined with the
ideas presented in this work to understand reasoning quantum mechanically.  For example, the thought vector space might ``factorize'' into
products of representations of Lie algebras.  Correspondingly, one might find e.g.\ translation works particularly well if one \emph{demands} the thought vector
space to be built up in this manner and goes on to use the quantum rule for
probability to interpret it.  

Then, ``thinking'' about some topic could actually best be represented by a matrix, with 
definitely having certain thoughts then being represented by thought \emph{eigen}vectors (the vector space becoming a Hilbert space).  The impossibility 
of holding at least certain beliefs simultaneously
(non-commutation of thoughts) would then emerge naturally.  The simplification that the superposition principle brings in organising possibilities and reducing the
dimensionality of the problem
would presumably
be very helpful.  Indeed, if the brain does actually work in something approximating this manner, evolution having selected for such efficiency, the brain could then be 
said to be behaving quantum mechanically, though not of course in the
sense in which this phrase might have been used before\footnote{For
  related points of view see the introductory text on quantum
  cognition \cite{bb2012} mentioned in the introduction and references
  therein.   Also note Ref.\ \cite{BUSEMEYER201753} which investigates
  ways in which it is possible for neural networks to implement such calculations.}.

Finally, the efficiency and other gains that a quantum approach bring might not be limited solely to the human brain.  It ought to be possible for artificial intelligence
to also reap the benefits.  Relating questions to one another should reduce the need for retraining (e.g.\ a new ``is it a guinea pig?'' question to some image classifier might just be a linear combination of the already-trained-for ``is it a cat?'' and ``is it a mouse?'' questions, requiring no further training).  The natural output of an 
analysis would be the state vector, implicitly containing
probabilistic answers to all askable questions.\footnote{One might further speculate on implications of a quantum viewpoint onto the
  very construction of approaches to machine learning.  For example, rather than combining and passing real
numbers from layer to layer of a network, perhaps one should use mini state vectors transforming under representations of a Lie algebra,
composing appropriately at each input stage and perhaps ``measured''
for each output.  Note that this would be distinct from using quantum computers to speed up training for current network designs.}    The answer to any such question 
would be one of a discrete set of values, with probabilities ensuring a caution or ``tolerance'' \cite{bron} built into any inference.  Given the complete classification of
Lie algebras, a certain commonality or universality \emph{between} problems might be exploitable also; once one has solved some problem matching on to say su(4), solving
another su(4) problem might be straightforward, and going on to solve a problem matching on to su(5)  might not be so difficult either.

\subsection{Counterpart to Unitary Evolution}

As a quantum system interacts, its state evolves with time, as we saw when discussing magnetic fields causing procession of the spin axis of a particle.  Thus,
before making a measurement, one can control the evolution of a system in a previously known state to make a given outcome more or less likely.  In everyday
life this might correspond to a restaurateur, knowing a particular client would like a glass of white wine say, drops hints about how nice the ros\'{e} is before asking
if the client would indeed like the ros\'{e} he or she is trying to clear.

Most people might naturally use low-dimensional representations for many issues.  Advertising and marketing might effectively correspond to unitary evolution before a measurement process such as a purchase decision or a vote.   Determining if such analogies exist and recognizing them if they do might lead one to more effective
public information campaigns for example.  The recent Brexit vote in the UK, see e.g.\ \cite{shipman,cummings}, might make an interesting case study for the first part
of this at least.

Given the option of providing a quantum description of the measurement
process in physics, it is of interest to consider parallels to this in
quantum inference.  The interaction term in the Hamiltonian, which
provides the quantum description by correlating one observable
with another, might correspond to a ``policy" for putting thoughts into
action.  For example, alternately applying to a state vector a term
coupling ``should I turn the
steering wheel clockwise?" to the question ``should I turn right?"
and a term coupling ``should I accelerate?" to ``is it clear in
front?" and so on might form the basis for
a self-driving car implementation.  Giving an option to steer clockwise
then ``measures'' a desire to turn right and an option to accelerate
``measures'' whether there is plenty of space in front.  Non-commutation
between the thoughts would then naturally express how for
example actions taken in response to one thought force other thoughts
to be reconsidered.

\section{Summary}

This work points out how the treatment of quantum-mechanical reasoning as presented by Dirac may
be applicable to situations in everyday life.  The representation theory of Lie algebras guides us in organising
both sets of questions and answers in an appropriate way.  Gleason's theorem gives us confidence that we
are extracting numerical probabilities from the formalism in an appropriate manner.  

As alluded to in Sec.\ \refsec{olagt} above, in Ref.\ \cite{dirac}, Dirac says of quantum mechanics that ``there remains an overall criticism \ldots\ that in departing from \ldots\ determinacy \ldots\ a great complication is introduced \ldots\ which is a highly undesirable feature. This complication is undeniable, but it is offset by a great simplification, provided by the general principle of superposition of states \ldots''.  For applications
to inference however, we consider \emph{both} a lack of determinacy \emph{and} a superposition principle to be desirable features, enabling responses to include uncertainty and allowing for
analysis reuse. 

It would be interesting to investigate how widely and how precisely these ideas actually represent the real life behaviour 
of people and organisations, whether they can aid us in advancing machine learning, and to see if such a structure is 
helpful in understanding thought and meaning in human and artificial intelligence.

\section*{Acknowledgements}

I thank Anthony Aguirre, Jeremy Butterfield, Adrian Kent, Antony
Lewis, Andrew Pontzen, Emmanuel Pothos, Harvey Reall and Neil Turok
for useful discussions and George Efstathiou, Emmanuel Pothos and Neil
Turok for comments on drafts of this paper.  This work was supported in part by an FQXi minigrant.

\bibliographystyle{unsrt}

\end{document}